\documentclass[conference]{IEEEtran}

\IEEEoverridecommandlockouts
\usepackage{cite}
\usepackage{amsmath,amssymb,amsfonts}
\usepackage{algorithm} % 用于 algorithm 环境
\usepackage{algpseudocode}
\usepackage{float}
\usepackage{graphicx}
\usepackage{textcomp}
\usepackage{xcolor}
\usepackage{url}
\usepackage{booktabs}
\usepackage{multirow}
\usepackage{threeparttable}
\usepackage[top=0.75in, bottom=1.25in, left=0.625in, right=0.625in]{geometry}

\def\BibTeX{{\rm B\kern-.05em{\sc i\kern-.025em b}\kern-.08em
    T\kern-.1667em\lower.7ex\hbox{E}\kern-.125emX}}
\begin{document}

\title{GTCN-G: A Residual Graph-Temporal Fusion Network for Imbalanced Intrusion Detection
}

\author{
Tianxiang Xu\textsuperscript{1$\dagger$}\thanks{$\dagger$~These authors contributed equally to this work},
Zhichao Wen\textsuperscript{2$\dagger$},
Xinyu Zhao \textsuperscript{3},
Qi Hu\textsuperscript{4},
Yan Li\textsuperscript{5},
Chang Liu\textsuperscript{6*}\thanks{*~Corresponding author: Chang Liu (cryyu1478@outlook.com)}
\\[0.5ex]
\textsuperscript{1}Peking University, Beijing, China\\
\textsuperscript{2}RWTH Aachen University, Aachen, Germany\\
\textsuperscript{3}University of Texas at Austin, Austin, USA\\
\textsuperscript{4}Northeastern University, Vancouver, Canada\\
\textsuperscript{5}Thales Group, Ottawa, Canada\\
\textsuperscript{6}Chinese Medical Information and Big Data Association, Beijing, China
}

\maketitle

% 直接添加首页脚注（\maketitle之后，正文之前）
\footnotetext{\small For any commercial use or derivative works, please contact the IEEE Copyrights Office at copyrights@ieee.org.}

\begin{abstract}
The escalating complexity of network threats and the inherent class imbalance in traffic data present formidable challenges for modern Intrusion Detection Systems (IDS). While Graph Neural Networks (GNNs) excel in modeling topological structures and Temporal Convolutional Networks (TCNs) are proficient in capturing time-series dependencies, a framework that synergistically integrates both while explicitly addressing data imbalance remains an open challenge. This paper introduces a novel deep learning framework, named Gated Temporal Convolutional Network and Graph (GTCN-G), engineered to overcome these limitations. Our model uniquely fuses a Gated TCN (G-TCN) for extracting hierarchical temporal features from network flows with a Graph Convolutional Network (GCN) designed to learn from the underlying graph structure.  The core innovation lies in the integration of a residual learning mechanism, implemented via a Graph Attention Network (GAT). This mechanism preserves original feature information through residual connections, which is critical for mitigating the class imbalance problem and enhancing detection sensitivity for rare malicious activities (minority classes). We conducted extensive experiments on two public benchmark datasets, UNSW-NB15 and ToN-IoT, to validate our approach. The empirical results demonstrate that the proposed GTCN-G model achieves state-of-the-art performance, significantly outperforming existing baseline models in both binary and multi-class classification tasks.
\end{abstract}


\section{Introduction}

As Internet technology becomes increasingly embedded in military, economic, scientific, and daily life, the threat posed by network attacks grows due to their complexity and variety~\cite{Teng2019}. High-profile incidents such as the WikiLeaks release of CIA documents and the spread of ransomware like WannaCry and Petya illustrate the urgency of effective network security. Intrusion Detection Systems (IDS) play a vital role in this defense, commonly using anomaly or misuse detection. Misuse detection is effective for known threats but struggles with unknown attacks, while anomaly detection offers adaptability at the cost of higher false alarms. The ever-expanding Internet scale, legacy security issues, and rise of Advanced Persistent Threats (APT) have driven the evolution of attack tactics. As shallow models fail to handle the sophistication of modern threats, deep learning has emerged as a powerful tool by building hierarchical feature representations that enhance detection accuracy and responsiveness.

Flow-based data, such as NetFlow~\cite{RFC3954}, is commonly used in IDS. A network flow is typically characterized by endpoint identifiers (IP address, L4 port, protocol) and enriched with features like packet count, bytes, and duration. These flows can be represented in graph form, where endpoints are nodes and flows are edges, making topological and feature-based information essential for identifying malicious traffic. Graph-based representations enhance the IDS's ability to capture complex relationships in network behavior. As network topologies and traffic patterns grow increasingly complex, graph structures combined with machine learning provide a scalable way to maintain detection efficacy.

\begin{figure}[H]
\centering
% 将此处替换为包含上面 SVG 的 .pdf 或 .svg 图像路径
\includegraphics[width=\linewidth]{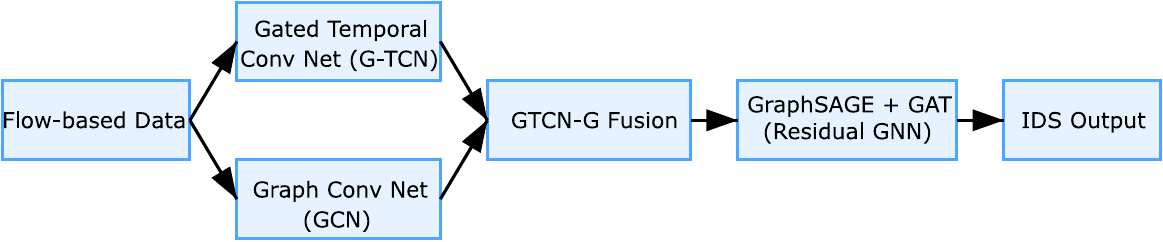}
\caption{Architecture of the proposed GTCN-G and Graph-SAGE-based intrusion detection method.}
\label{fig:gtcn-graphsage-architecture}
\end{figure}

Recent research has demonstrated the significant potential of deep learning in addressing the challenges of large-scale and high-dimensional IDS data~\cite{Pozi2015}. Early deep learning models, including RNNs~\cite{Roy2017}, deep autoencoders~\cite{Yang2007}, DBNs~\cite{Alom2016}, and various hybrid models~\cite{Tan2016}, have shown improved detection accuracy, albeit sometimes at the cost of computational efficiency. More recently, Graph Neural Networks (GNNs) have emerged as a particularly effective paradigm by modeling network traffic as a graph to capture intricate topological relationships. While these GNN-based approaches have advanced the field, many tend to focus on static graph structures, often underexploiting the rich temporal dynamics inherent in network flows. Furthermore, mitigating the severe class imbalance that characterizes most real-world traffic remains a critical challenge. To address these gaps, this paper proposes a novel method, the Gated Temporal Convolution Network and Graph (GTCN-G). Our approach synergistically integrates a Gated Temporal Convolution Network (G-TCN) to capture temporal dependencies with a Graph Convolutional Network (GCN) to learn structural patterns. Crucially, we incorporate a residual learning mechanism via a Graph Attention Network (GAT) to preserve original features, thereby enhancing detection robustness for minority (malicious) classes. The architecture, illustrated in Figure~\ref{fig:gtcn-graphsage-architecture}, processes data through parallel G-TCN and GCN modules, fuses their outputs, and refines the representation to produce a final classification.

The main contributions of this paper are as follows:

\begin{enumerate}
\item A novel intrusion detection method, GTCN-G, is proposed. It combines a Gated Temporal Convolution Network (G-TCN) and a Graph Convolutional Network (GCN) to concurrently utilize both temporal edge features and static network topology.
\item A residual learning mechanism is introduced for the graph-based component. This mechanism employs a Graph Attention Network (GAT) within a GraphSAGE-inspired framework to preserve original information and improve the detection of minority classes.
\item A specialized gated temporal convolution network model is developed to hierarchically capture temporal features from network traffic, specifically for the task of intrusion detection.
\item Extensive experiments on the UNSW-NB15 and ToN-IoT datasets are presented, which demonstrate that the proposed GTCN-G model achieves superior detection performance compared to established baselines.
\end{enumerate}

\section{Related Work}

Network intrusion detection systems have evolved from rule-based approaches to sophisticated machine learning frameworks. Traditional methods including threshold analysis, statistical modeling, and signature-based detection demonstrate fundamental limitations when processing large-scale network traffic due to their dependence on historical observations and expert domain knowledge. These approaches frequently exhibit elevated false positive rates and diminished detection capabilities as network conditions evolve, necessitating more adaptive detection mechanisms.

\subsection{Machine Learning and Deep Learning Approaches}

Early machine learning approaches include Li et al.~\cite{li2018ai}'s dual-phase methodology incorporating the Bat Algorithm for feature selection in SDNs, Gao et al.~\cite{gao2019ensemble}'s ensemble learning with decision tree optimization, Yulianto et al.~\cite{yulianto2019adaboost}'s AdaBoost integration with feature selection mechanisms, Jan et al.~\cite{jan2019lightweight}'s lightweight SVM frameworks for IoT environments, and Abubakar et al.~\cite{abubakar2017mlsdsn}'s dual IDS architectures for SDN networks. However, shallow learning methods encounter limitations including computational overhead, overfitting susceptibility, and reliance on manual feature engineering that constrains generalizability across diverse attack vectors.

Deep learning architectures address these limitations through autonomous feature extraction from raw network data. Chen et al.~\cite{Chen2020} developed the RULA-IDS framework with global attention mechanisms for capturing temporal and spatial traffic characteristics. Chkirbene et al.~\cite{Chkirbene2020} introduced adaptive weighted classification methodologies for imbalanced datasets, while Wang et al.~\cite{Wang2020} demonstrated Stack Shrinkage Automatic Encoder (SCAE) integration with SVM for multi-dimensional traffic analysis. For imbalanced scenarios, Elsayed et al.~\cite{elsayed2020traffic} pioneered autoencoder-LSTM hybrid architectures in unsupervised contexts, Tang et al.~\cite{tang2018rnn} applied GRU-RNN for DDoS prevention in SDN environments, Farahnakian et al.~\cite{farahnakian2018dae} employed Deep Automatic Encoder with Softmax classification, and Al-Qatf et al.~\cite{alqatf2018sparse} combined sparse autoencoders with SVM for enhanced dimensionality reduction. These approaches consistently demonstrate superior accuracy while reducing error rates~\cite{Ayachi2020} and enabling automated identification of attack attributes and emerging threat patterns.

\subsection{Graph Neural Networks in Network Security}

Graph-based deep learning represents a rapidly expanding research domain with substantial implications for network security. The foundational Graph Attention Network (GAT)~\cite{velickovic2017graph} has been extensively adopted in communication networks, distinguished by sophisticated attention mechanisms and multi-head attention architectures that enhance learning stability. Yu et al.~\cite{yu2020mstnn} developed the Multiscale Spatial-Temporal Graph Neural Network (MSTNN), employing dynamic attention mechanisms for capturing temporal variations in spatial node correlations.

Recent advances demonstrate growing adoption of GNN architectures for intrusion detection tasks. Wan et al.~\cite{wan2021glad} proposed GLAD-PAW for log file anomaly detection with location-aware weighted graph attention layers. Chen et al.~\cite{chen2021attack} developed Alert-GCN to correlate security alarms within attack sequences through node classification. Zhang et al.~\cite{zhang2020botnet} utilized GCN architectures for botnet identification by integrating connection patterns from malicious traffic with network topologies. Contemporary contributions include the VAE-GNN framework~\cite{vae2024adma} for enhanced intrusion detection, heterogeneous GNN architectures with express edges~\cite{hetgnn2024icnc} for cyber-physical systems, interpretable provenance-based detection using GNNExplainer~\cite{gnnexpids2024scisec}, the comprehensive GNN-IDS framework~\cite{gnnids2024ares}, memory-replay approaches~\cite{memreplay2025icoin}, efficient network representation methodologies~\cite{efficientgnn2023acns}, and cyber-physical GNN implementations for smart power grids~\cite{cyberphy2023smartgrid}.

\subsection{Temporal Convolutional Networks and Hybrid Integration}

Temporal Convolutional Networks (TCN) have demonstrated exceptional performance in time series analysis tasks, making them particularly relevant for network traffic pattern recognition. Recent research has explored TCN applications in intrusion detection, with Chen et al.~\cite{chen2021efficient} developing an efficient network intrusion detection model, Lopes et al.~\cite{lopes2023network} proposing temporal convolutional model-based detection, Derhab et al.~\cite{derhab2020intrusion} applying TCN for IoT-based intrusion detection, and de Araujo-Filho et al.~\cite{de2023unsupervised} developing unsupervised GAN-based systems using TCN and self-attention. However, systematic integration of TCN with graph-based approaches remains underexplored. The proposed GTCN-G approach addresses this gap by combining Gated Temporal Convolution Networks with Graph Convolutional Networks, leveraging both temporal feature extraction and structural pattern recognition capabilities to provide a comprehensive framework that advances beyond existing approaches focusing solely on temporal or spatial characteristics.

\section{Method}
\subsection{Graph Construction}
\begin{figure}
\centering
\includegraphics[width=0.4\textwidth]{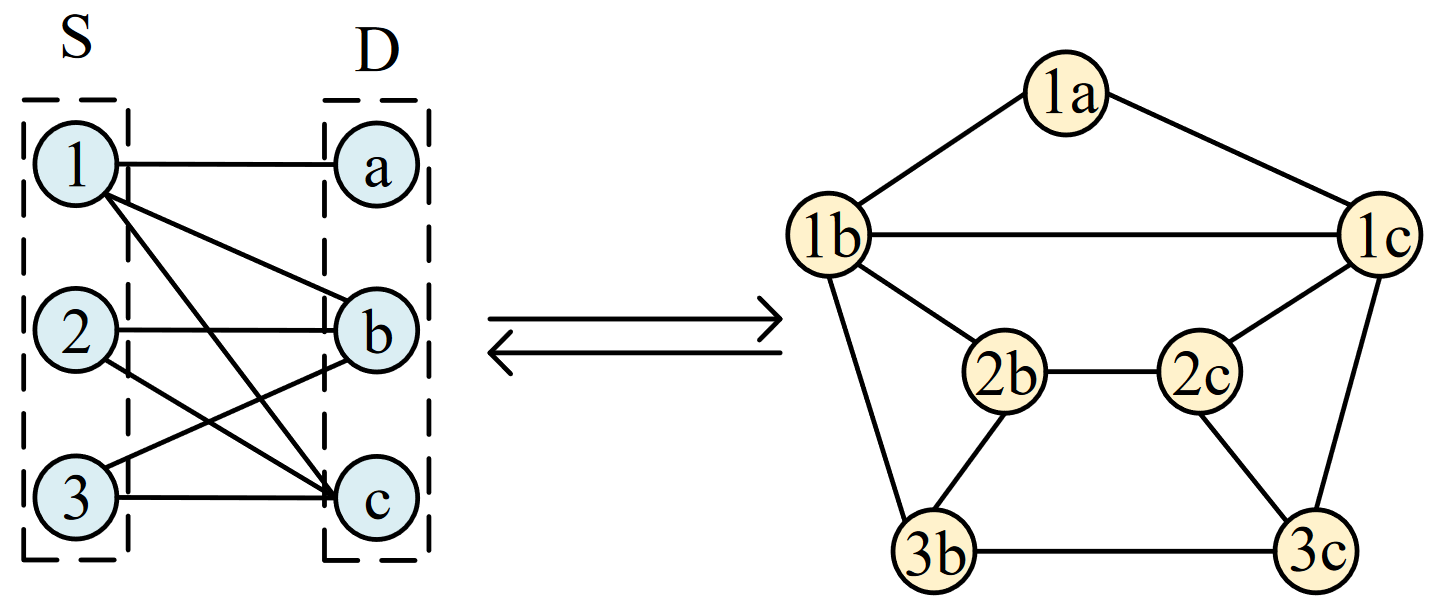}
\caption{Bipartite Graph to Line Graph Transformation for Network Flow Edge Classification.} \label{fig1}
\end{figure}

Three components make up network flow data: traffic characteristics, such as time, transaction bytes, and transmitted packet size; and the source and destination addresses. To create a graph, the source and destination identifiers are used as the endpoints of a network flow. To model this, we define each unique endpoint (i.e., an IP address and port number combination) as a \textbf{node} in the graph. A network flow between a source and a destination then constitutes an \textbf{edge} connecting the corresponding nodes. The remaining traffic data are used as edge features. The intrusion detection problem is thus framed as an edge classification task. To formally distinguish the roles of traffic initiators and receivers, this paper constructs a bipartite graph $G(S,D;E)$. In this graph, $S$ represents the set of all source nodes, $D$ represents the set of all destination nodes, and $E$ is the edge set representing the flows between them. Therefore, the bipartite graph can be converted to its corresponding line graph, in which the original edges become the new nodes. As a result, the problem is changed to a node classification task, as seen in Fig.~\ref{fig1}. The transformation shown in Fig.~\ref{fig1} illustrates how a node in the line graph (right) corresponds to an edge in the original bipartite graph (left).

In addition, if $|S|<|D|$~(resp.\ $|S|>|D|$), virtual nodes are created to increase the size of the source (resp.\ destination) node set to match that of the target (resp.\ source) node set. The reasons are as follows:

(1) For GAT-based models, using the complete neighborhood of high-degree nodes may prevent the line graph's adjacency list from loading into memory. While the number of edges $|E|$ in the bipartite graph remains unchanged, this padding step increases the node count and thus lowers the average node degree. This is beneficial because the number of edges in the resulting line graph, which is given by $\sum_{i \in S \cup D} (d_i - 1)d_i/2$, depends on the node degrees $d_i$. In this manner, the memory required for the line graph can be decreased.

(2) Second, introducing virtual nodes adds a degree of random mapping, which helps to avoid potential biases where specific source nodes might disproportionately influence traffic classification.

\begin{figure}
\includegraphics[width=0.45\textwidth]{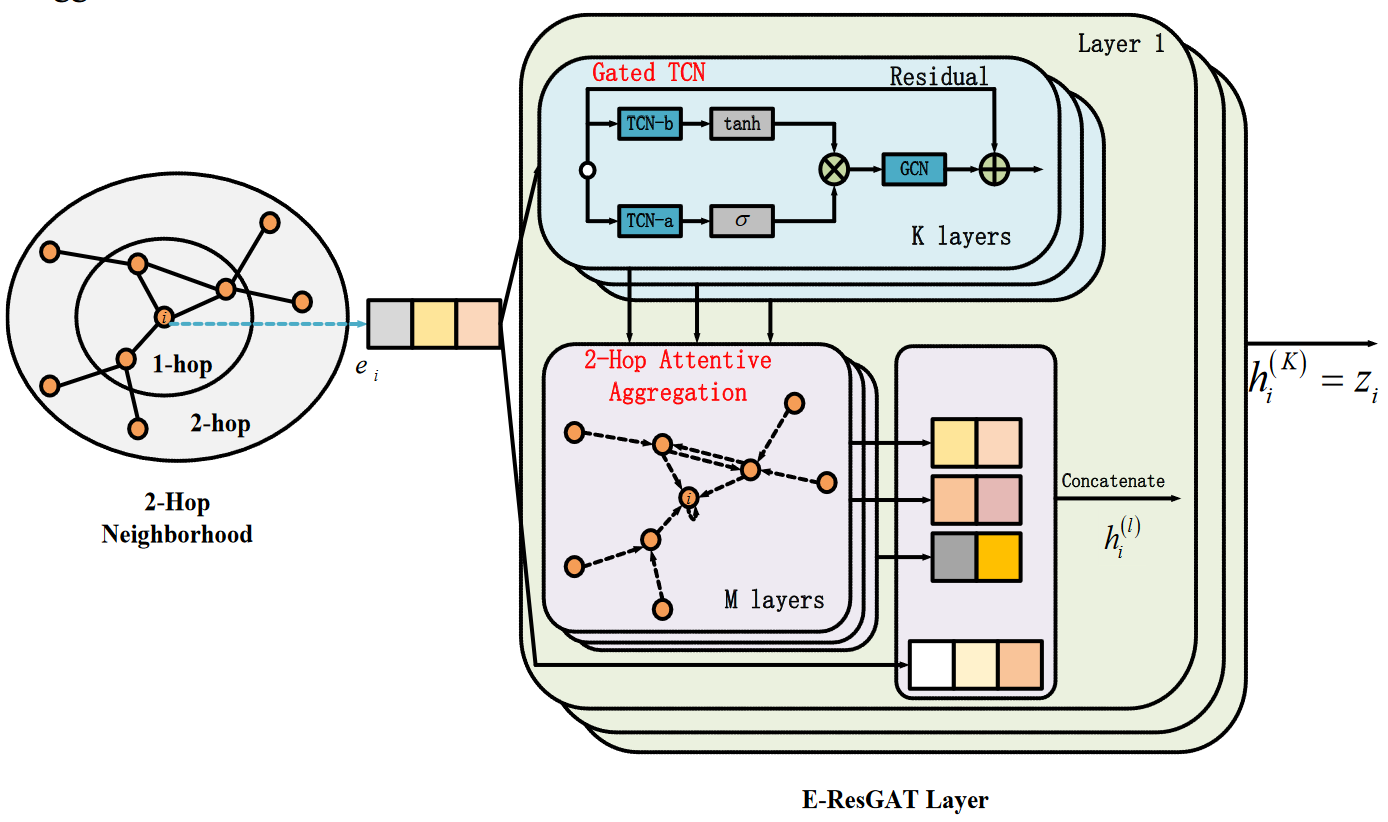}
\caption{GTCN-G network structure diagram.} \label{fig2}
\end{figure}

\subsection{Graph-SAGE Framework}

To assist edge classification, we utilize a method that enhances the E-GraphSAGE algorithm~\cite{lo2021egraphsage}. E-GraphSAGE's main concept is to run the GraphSAGE algorithm~\cite{hamilton2017inductive} on both of an edge's endpoints before concatenating the resulting node embeddings to form an edge representation. The main contribution of this work, within this framework, is a modification to the final embedding step, aiming to overcome the potential loss of original edge feature information. The framework consists of two main components: sampling and aggregation. Due to memory limitations, this work implements a mini-batch version of the E-GraphSAGE algorithm, which is referred to as E-GraphSAGE-M in the experiments. Algorithm 1 and Algorithm 2 illustrate the sampling and aggregation procedures.

Sampling: In the GraphSAGE algorithm, a fixed-size set of neighbors is uniformly sampled for each node. For a given node $v$, its full neighborhood $N(v)$ is the set of all adjacent nodes, i.e., $N(v) = \{u \in V \mid (u,v) \in E\}$. The sampling process then creates a fixed-size subset of $N(v)$. For each edge batch $B$, different uniform samples are drawn from the neighborhood of each relevant node at each layer of aggregation.

Without such sampling, the memory and runtime of a single batch can be unpredictable, scaling with $O(|V|)$ in the worst case. Therefore, before aggregation, we sample the batch's 2-hop neighborhood.

\vspace{3mm} %
\begin{algorithm}[H]
    \caption{Graph-SAGE Minibatch Sampling}
    \label{alg:sampling_corrected}
    \small
    \begin{algorithmic}[1]
        \Require 
        \Statex Graph $G(V,E)$;
        \Statex Edge minibatch $B$;
        \Statex Node set of the batch, $V(B) = \{v \mid v \text{ is an endpoint of an edge in } B\}$;
        \Statex Search depth $K$;
        \Statex Neighborhood sampling function $N: v \rightarrow 2^V$;
        \Ensure 
        \Statex A K-hop neighborhood of the batch, represented by the final edge set $B^0$;

        \State $B^K \gets B$ \Comment{Initialize with the minibatch}
        \For{$k \gets K$ \textbf{to} $1$ \textbf{step} $-1$} \Comment{Iterate backwards from layer K to 1}
            \State $B^{k-1} \gets B^k$
            \ForAll{$v \in V(B^k)$} \Comment{$V(B^k)$ are the nodes in the current edge set}
                \State $B^{k-1} \gets B^{k-1} \cup \{ (u, v) \mid u \in N(v) \}$
            \EndFor  % <--- 已修正
        \EndFor  % <--- 已修正
        \State \Return $B^0$ \Comment{Return the K-hop neighborhood}
    \end{algorithmic}
\end{algorithm}

\textbf{Aggregation}: Following sampling, the method iteratively aggregates features from neighboring nodes layer by layer. The aggregation process for a single layer is as follows: given input edge features $h_{uv}^{k-1}$, we first aggregate them for a target node $v$:

\begin{equation}
\mathbf{h}_{N(v)}^k = \operatorname{AGG}_k \left( \left\{ \mathbf{h}_u^{k-1} : u \in N(v) \right\} \right)
\end{equation}

In this article, mean aggregation is used to compute the average of features in the sampled neighborhood. Then, the aggregated information $h_{N_v}^k$ is combined with the node's embedding from the previous layer $h_v^{k-1}$, passed through a trainable weight matrix $W_k$, and activated by  $\sigma$ to generate the new node embedding.

\begin{equation}
\mathbf{h}_v^k = \sigma \left( \mathbf{W}^k \left[ \mathbf{h}_v^{k-1} \Vert \mathbf{h}_{N(v)}^k \right] \right)
\end{equation}

Where $\|$ stands for concatenation. Since nodes in the initial graph have no features, they are initialized with an all-ones vector $h_v^0$, whose dimension matches the edge features. After $K$ layers of aggregation, the standard edge embedding is the concatenation of its endpoints' final node embeddings:

\begin{equation}
\mathbf{z}_{uv} = \mathbf{h}_u^K \Vert \mathbf{h}_v^K, \quad \forall uv \in B
\end{equation}

However, the iterative aggregation process might dilute the initial edge features. To overcome this, our modification enhances the final edge embedding by concatenating the original edge feature $e_{uv}$ as well:

\begin{equation}
\mathbf{z}_{uv} = \mathbf{h}_u^K \Vert \mathbf{h}_v^K \Vert \mathbf{e}_{uv}, \quad \forall uv \in B
\end{equation}

This $z_{uv}$ represents the final edge embedding from our modified Graph-SAGE framework.

\vspace{5mm}
\begin{algorithm}
\caption{Graph-SAGE minibatch aggregation}
\small
\begin{algorithmic}[1]
\Require Graph $G(V,E)$;
\Statex \quad edge minibatch $B$; 2-hop neighborhood $B^0$;
\Statex \quad node set $V(B)=\{v: v \text{ is an end point of } uv \in B\}$;
\Statex \quad input edge features $\{e_{uv}, \forall uv \in B\}$;
\Statex \quad input node features $\mathbf{x}_v = 1$;
\Statex \quad depth or layer $K$; non-linearity $\sigma$;
\Statex \quad weight matrices $W^k, \forall k \in \{1,...,K\}$;
\Statex \quad differentiable aggregator functions AGG$_k$;
\Statex \quad neighborhood sampling functions $N_v : v \rightarrow 2^v$;
\Ensure Vector representation $z_{uv}$, $\forall uv \in B$
\State $h_{uv}^0 \leftarrow e_{uv}, \forall uv \in B^0$;
\State $h_v^0 \leftarrow \mathbf{x}_v, \forall v \in V(B^0)$;
\For{k=1,\ldots,K}
    \For{$v \in V(B^k)$}
        \State $h_v^k = \sigma\left(W^k \bullet \left[h_v^{k-1} \| h_{N_v}^k\right]\right)$;
    \EndFor
\EndFor
\State $z_{uv} = \|\left(\{h_u^K, h_v^K, e_{uv}\}\right), \forall uv \in B$;
\end{algorithmic}
\end{algorithm}

\subsection{The Proposed GTCN-G Model Architecture}
A limitation of the baseline GraphSAGE framework is that it treats every adjacent edge uniformly. In network security contexts, however, learning weighted representations of neighbors is crucial for identifying subtle attack patterns. Therefore, to achieve this weighted aggregation, our proposed GTCN-G model integrates a Gated Temporal Convolution Network (Gated TCN), a Graph Convolutional Network (GCN), and a Graph Attention Network (GAT). A key architectural difference from the baseline approach is that GTCN-G operates directly on the line graph $G_L(V_L, E_L)$, rather than on the original bipartite graph. This allows sampling from a richer neighborhood that captures higher-order relationships between flows.

\begin{figure*}[t]
\centering
% 将此处替换为包含上面 SVG 的 .pdf 或 .svg 图像路径
\includegraphics[width=0.85\linewidth]{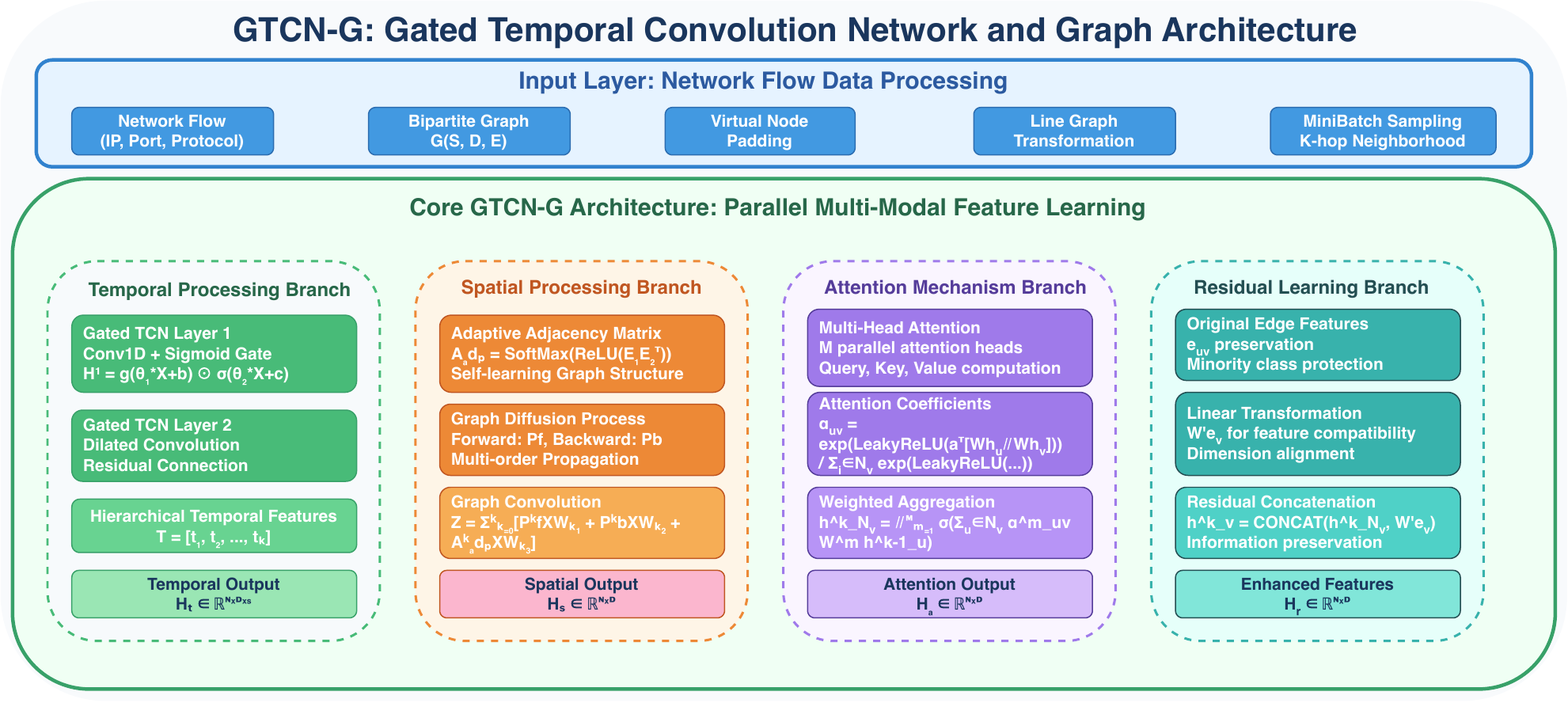}
\caption{The detailed architecture of the proposed GTCN-G model. Input data, represented as a line graph, is processed through four parallel feature-learning branches: (a) a \textbf{Temporal Branch} using Gated TCNs to capture time-series dependencies; (b) a \textbf{Spatial Branch} using an adaptive graph convolution to learn topological patterns; (c) an \textbf{Attention Branch} employing multi-head attention for weighted neighbor aggregation; and (d) a \textbf{Residual Branch} that preserves original node features to combat class imbalance. The outputs from these branches are then fused for final classification.}
\label{fig:framework}
\end{figure*}

The general structure of the proposed GTCN-G model is depicted in Figure~\ref{fig2}. The model's aggregation process is fundamentally more advanced than the baseline. Instead of simple averaging, GTCN-G leverages its GAT component for attention-based aggregation, while simultaneously using the G-TCN and GCN to extract temporal and spatial features. Furthermore, GTCN-G incorporates a residual connection that concatenates aggregated neighborhood features with the node's transformed original features. This residual learning mechanism allows GTCN-G to build deeper models than a standard GAT, preventing information loss and progressively refining node embeddings. To provide a more detailed illustration, Figure~\ref{fig:framework} presents the complete architecture, showing how these components are organized into four parallel processing branches for comprehensive feature learning.

\subsection{Core Components and Mechanisms}
The GTCN-G model is composed of several key components designed to capture temporal, structural, and relational features from network traffic.

\subsubsection{Gated Temporal Convolutional Network (G-TCN)}
To effectively control the information flow through the temporal feature extraction layers, we employ a Gated TCN. This unit uses a learned sigmoid gate to modulate the output of a standard 1D convolution. For an input of $N$ sequences with length $S$ and $D$ features each ($X \in \mathbb{R}^{N \times S \times D}$), the output $H$ is:

\begin{equation}
    H = g(\theta_1 * X + b) \odot \sigma(\theta_2 * X + c)
\end{equation}

where $*$ represents the 1D convolution operator, while $\{\theta_1, b, \theta_2, c\}$ are its learnable parameters (filters and biases). The function $g(\cdot)$ is a primary activation like Tanh, and $\sigma(\cdot)$ is the sigmoid gate.

% \subsubsection{Gated Temporal Convolutional Network (G-TCN)}
% To effectively control the information flow through the temporal feature extraction layers, we employ a Gated TCN. This unit uses a learned sigmoid gate to modulate the output of a standard 1D convolution, determining the proportion of information passed to the next layer. For an input sequence $X \in \mathbb{R}^{N \times D \times S}$, the output $H$ is:
% \begin{equation}
%     H = g(\theta_1 * X + b) \odot \sigma(\theta_2 * X + c)
% \end{equation}
% where $g(\cdot)$ is an activation function (e.g., Tanh) and $\sigma(\cdot)$ is the sigmoid gate.

\subsubsection{Adaptive Graph Convolution}

Our graph convolution module captures spatial dependencies by extending the standard GCN framework. It integrates a graph diffusion process with a self-learning adaptive adjacency matrix ($A_{adp}$). The adaptive matrix learns latent structural relationships from two learnable node embedding matrices, $E_1$ and $E_2$:

\begin{equation}
    A_{adp} = \operatorname{SoftMax}\left( \operatorname{ReLU}\left(E_1 E_2^T\right) \right)
\end{equation}

The final composite graph convolution layer combines graph diffusion (using forward $P_f$ and backward $P_b$ transition matrices, derived from the graph adjacency) with the adaptive matrix to form a comprehensive feature representation:

\begin{equation}
    Z = \sum_{k=0}^{K} \left( P_f^k X W_{k1} + P_b^k X W_{k2} + A_{adp}^k X W_{k3} \right)
\end{equation}

\subsubsection{Attention-based Aggregation with Residuals}

GTCN-G utilizes a multi-head attention mechanism, parameterized by a weight matrix $W$ and an attention vector $\mathbf{a}$, to perform weighted aggregation. The attention coefficient $\alpha_{uv}$ between a node $v$ and its neighbor $u$ is computed as:

\begin{equation}
    \alpha_{uv} = \frac{\exp(\operatorname{LeakyReLU}(\mathbf{a}^T [W\mathbf{h}_u \Vert W\mathbf{h}_v]))}{\sum_{i \in N(v)} \exp(\operatorname{LeakyReLU}(\mathbf{a}^T [W\mathbf{h}_i \Vert W\mathbf{h}_v]))}
\end{equation}

A key innovation in our aggregation is the use of a residual connection. The outputs of the $M$ attention heads are aggregated to form a neighborhood vector $\mathbf{h}_{\text{agg}}^k$. This vector is then concatenated with a linear transformation of the node's original feature vector $\mathbf{e}_v$.

\begin{equation}
    \mathbf{h}_{\text{agg}}^k = \operatorname{CONCAT}_{m=1}^M \left( \sigma \left( \sum_{u \in N(v)} \alpha_{uv}^m W^m \mathbf{h}_{u}^{k-1} \right) \right)
\end{equation}

\begin{equation}
    \mathbf{h}_v^k = \operatorname{CONCAT}\left(\mathbf{h}_{\text{agg}}^k, W'\mathbf{e}_v\right)
\end{equation}

This mechanism ensures that a node's intrinsic features are preserved, which is critical for preventing performance degradation on imbalanced datasets where a node's neighbors may predominantly belong to the majority class.

\section{Experiment}  
\subsection{Datasets}
This study evaluates the proposed model on two public benchmark datasets: UNSW-NB15~\cite{moustafa2015unsw} and ToN-IoT~\cite{alsaedi2020toniot}. For the UNSW-NB15 dataset, the full set of provided instances is used, while for ToN-IoT, we utilize the official pre-defined training and testing splits.

\textbf{UNSW-NB15:} Produced by the Australian Cyber Security Centre (ACCS) in 2015, this dataset was generated by simulating attacks in a laboratory environment based on real-world threat intelligence from sources such as CVE and BID. It includes nine distinct attack categories: DoS, Exploits, Generic, Shellcode, Reconnaissance, Backdoor, Worms, Analysis, and Fuzzers.

\textbf{ToN-IoT:} Created in 2019, the ToN-IoT dataset provides a large-scale and diverse collection of data from an Internet of Things (IoT) and Industrial IoT (IIoT) testbed at UNSW Canberra. It contains not only network traffic but also operating system logs and telemetry data from various services.

Table~\ref{tab:datasets} provides a high-level overview of the two datasets. The detailed class distributions are presented in Table~\ref{tab:UNSW_Distribution} and Table~\ref{tab:ToN_IoT_Distribution}. A key characteristic evident in both datasets is a severe class imbalance. Each dataset is dominated by a single majority class (Normal) and features multiple minority classes representing various attacks. This distribution, where the majority class representation ranges from 96.83\% down to 65.07\%, poses a significant challenge for training robust multi-class classification models.

\begin{table}[H]
\centering
\caption{Overview of the Experimental Datasets.}
\label{tab:datasets}
\begin{tabular}{lccccc}
\toprule
Dataset & Examples & Normal (\%) & Classes & Features \\ 
\midrule
UNSW-NB15 & 700001 & 96.83 & 10 & 43 \\
ToN-IoT & 461043 & 65.07 & 10 & 39 \\
\bottomrule
\end{tabular}
\end{table}

\begin{table}[H]
\centering
\caption{Class distribution of UNSW-NB15 dataset.}
\label{tab:UNSW_Distribution}
\begin{tabular}{lc*{4}{c}}
\toprule
Dataset & \multicolumn{5}{c}{Classes (names and \%)} \\
\cmidrule(lr){2-6}
 & Normal & Exploits & Recon. & DoS & Generic \\
\midrule
UNSW-NB15 & 96.83 & 0.773 & 0.251 & 0.167 & 1.07 \\
\addlinespace
\midrule
 & Shellcode & Fuzzers & Worms & Backd. & Analysis \\
\midrule
UNSW-NB15 & 0.032 & 0.722 & 0.003 & 0.076 & 0.075 \\
\bottomrule
\end{tabular}
\end{table}

\begin{table}[H]
\centering
\caption{Class distribution of ToN-IoT dataset.}
\label{tab:ToN_IoT_Distribution}
\begin{tabular}{lc*{4}{c}}
\toprule
Dataset & \multicolumn{5}{c}{Classes (names and \%)} \\
\cmidrule(lr){2-6}
 & Normal & Scanning & DoS & Inject & DDoS \\
\midrule
ToN-IoT & 65.07 & 4.34 & 4.34 & 4.34 & 4.34 \\
\addlinespace
\midrule
 & Password & XSS & Ransomw. & Backd. & MITM \\
\midrule
ToN-IoT & 4.34 & 4.34 & 4.34 & 4.34 & 0.22 \\
\bottomrule
\end{tabular}
\end{table}

\subsection{Evaluation Metrics}
Given the significant class imbalance in the selected datasets, standard accuracy is not a reliable performance indicator. Therefore, we primarily use the F1-score for evaluation, as it provides a robust measure by calculating the harmonic mean of Precision and Recall.
\begin{equation}
    F1\text{-}score = 2 \times \left( \frac{Precision \times Recall}{Precision + Recall} \right)
\end{equation}

For multi-class classification, we report the weighted F1-score. This metric calculates the F1-score for each class independently and then computes an average, weighted by the number of true instances for each class (the support). This approach accounts for label imbalance and provides a more comprehensive assessment of the model's overall performance.

The component metrics, Precision and Recall, are defined as follows:
\begin{equation}
Precision = \frac{TP}{TP + FP}
\end{equation}

\begin{equation}
Recall = \frac{TP}{TP + FN}
\end{equation}
where the terms are defined as: True Positive (TP) is the number of attack instances correctly classified as attacks; True Negative (TN) is the number of normal instances correctly classified as normal; False Positive (FP) is the number of normal instances incorrectly classified as attacks; and False Negative (FN) is the number of attack instances incorrectly classified as normal.

\begin{figure}[H]
    \centering
    \includegraphics[width=0.3\paperwidth]{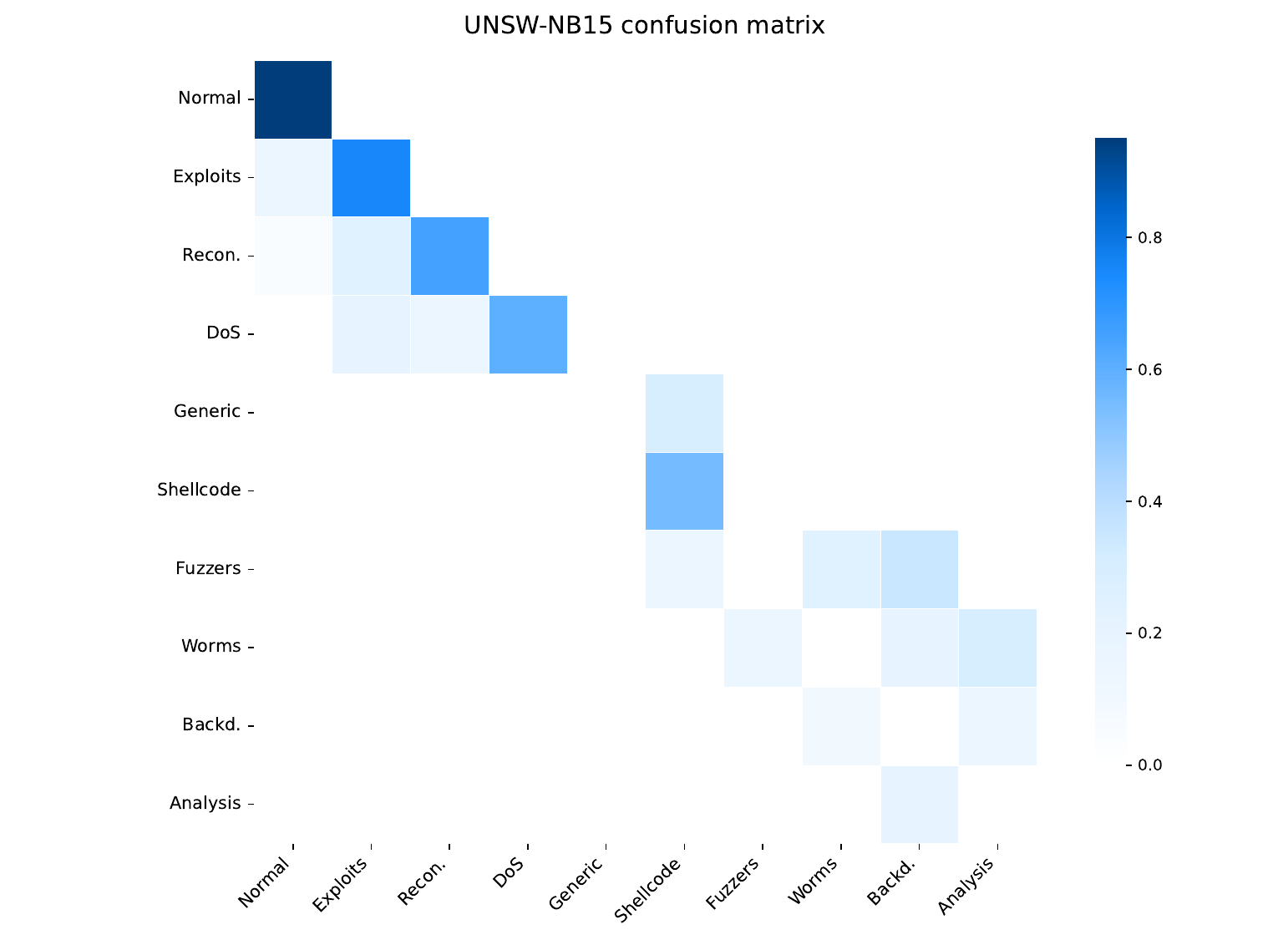}
    \caption{Multi-class confusion matrix for the UNSW-NB15 dataset.}
    \label{fig4} % 图片标签，用于后续引用
\end{figure}

\begin{figure}[H]
    \centering
    \includegraphics[width=0.3\paperwidth]{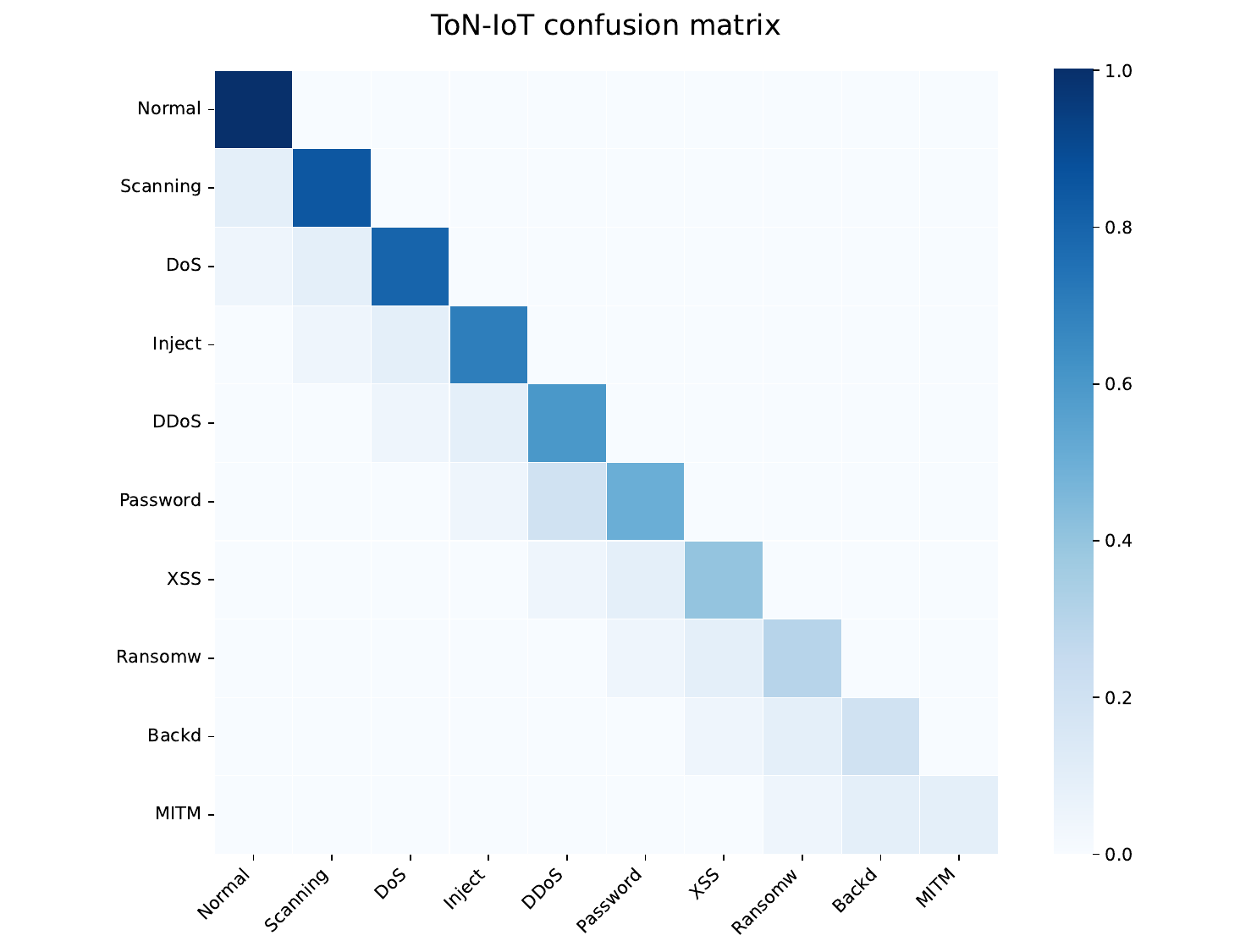}
    \caption{Multi-class confusion matrix for the ToN-IoT dataset.}
    \label{fig5} % 图片标签，用于后续引用
\end{figure}

\subsection{Experimental Settings}
For a fair comparison, all models were trained and evaluated under consistent conditions. The datasets were split into training, validation, and test sets with a ratio of 5:2:3. All models were trained for 10 epochs using a minibatch size of 500. The learning rates were optimized on the validation set, with the final rates set to 0.007 for the UNSW-NB15 dataset and 0.01 for the ToN-IoT dataset.

\subsubsection{Baseline Models}
We compare our proposed model against several baselines:
\begin{itemize}
    \item E-GraphSAGE-M: This is our implementation of the E-GraphSAGE framework with mini-batching. It is configured as a two-layer model using a mean aggregator, ReLU activation, and a 2-hop neighborhood sampling strategy with a sample size of 8 per hop.
    \item GAT: A standard Graph Attention Network model is used as a baseline to specifically ablate the effect of the residual connections in our proposed model.
\end{itemize}

\subsubsection{Proposed GTCN-G Model}
Our proposed GTCN-G model is configured as a deeper network where the optimal number of layers was determined via hyperparameter tuning on the validation set. Each layer employs a 6-head attention mechanism. To prevent overfitting, we apply dropout to the attention coefficients with a rate of 0.5.

\begin{table}[H]
\centering
\caption{F1-score Performance on Two Datasets}
\label{tab:F1}
\begin{tabular}{lc@{\hspace{2em}}cc}
\toprule
\textbf{Dataset} & \textbf{Algorithm} & \multicolumn{2}{c}{\textbf{F1-score}} \\
\cmidrule(l){3-4}
 &  & \textbf{Binary} & \textbf{Multi-Class} \\
\midrule
UNSW-NB15 & E-GraphSAGE   & 0.9023 & 0.8756 \\
 & E-GraphSAGE-M & 0.9234 & 0.8934 \\
 & GAT           & 0.9456 & 0.9178 \\
 & \textbf{GTCN-G}      & \textbf{0.9689} & \textbf{0.9512} \\
\addlinespace
ToN-IoT   & E-GraphSAGE   & 0.9287 & 0.9001 \\
 & E-GraphSAGE-M & 0.9434 & 0.9167 \\
 & GAT           & 0.9612 & 0.9389 \\
 & \textbf{GTCN-G}      & \textbf{0.9834} & \textbf{0.9698} \\
\bottomrule
\end{tabular}
\end{table}

\subsection{Experimental Analysis}
We evaluate our model on both binary (Normal vs. Attack) and multi-class classification tasks across the two datasets. The overall performance, measured by the weighted F1-score, is presented in Table~\ref{tab:F1}.

\subsubsection{Overall Performance Comparison}
The results in Table~\ref{tab:F1} show our proposed GTCN-G model achieves state-of-the-art performance, significantly outperforming all baseline models across both datasets and tasks. A comparative analysis of the baselines reveals a clear performance progression that validates our design choices. The superior performance of GAT over E-GraphSAGE-M confirms the efficacy of its attention mechanism for learning weighted neighbor importance, as opposed to simple mean aggregation. The final and most substantial performance leap from GAT to our GTCN-G model then demonstrates the effectiveness of our architecture's core innovations: 1) the synergistic fusion of temporal features (via G-TCN) with graph structural information, and 2) the inclusion of residual connections, which enhances the model's ability to learn from imbalanced classes by preserving crucial features of minority attack types.

% \subsubsection{Overall Performance Comparison}
% The results in Table~\ref{tab:F1} demonstrate that our proposed GTCN-G model achieves state-of-the-art performance, significantly outperforming all baseline models across both datasets and tasks.

% A comparative analysis of the baselines provides key insights. Firstly, the performance of E-GraphSAGE-M surpasses the standard E-GraphSAGE, indicating that our modification to retain original edge features is beneficial. Secondly, GAT consistently outperforms E-GraphSAGE-M, which highlights the efficacy of the attention mechanism for learning weighted edge importance in intrusion detection tasks, as opposed to simple mean aggregation.

% Most importantly, the substantial performance gap between GTCN-G and GAT validates the effectiveness of our proposed architecture. This superiority is attributed to two main factors: 1) the synergistic fusion of temporal features (via G-TCN) with graph structural information, and 2) the inclusion of residual connections, which enhances the model's ability to learn from imbalanced classes by preserving crucial features of minority attack types.

\subsubsection{Per-Class Analysis with Confusion Matrices}
To further analyze the model's classification capabilities, especially for minority classes, we present the confusion matrices for the multi-class task on both datasets in Figure~\ref{fig4} and Figure~\ref{fig5}.

As shown in Figure~\ref{fig4} for the UNSW-NB15 dataset, the GTCN-G model yields a remarkably clean diagonal. This indicates high precision and recall across almost all categories. Notably, the model successfully identifies extremely rare attack types like ``Worms'' (0.003\%) and ``Shellcode'' (0.032\%), which are often misclassified by other methods. This demonstrates the model's high sensitivity to minority classes.

Similarly, the confusion matrix for the ToN-IoT dataset, shown in Figure~\ref{fig5}, confirms the model's robustness. GTCN-G effectively distinguishes between various attack types that often share similar traffic patterns, such as ``DoS'', ``DDoS'', and ``Scanning''. The near-perfect diagonal underscores the model's effectiveness on a more balanced, yet still challenging, multi-class IoT dataset. This detailed per-class analysis provides strong evidence that the architectural choices in GTCN-G, particularly the residual mechanism, directly contribute to its superior and well-balanced classification performance.

\section{Conclusion}
This paper addressed the dual challenges of detecting sophisticated network attacks and handling the severe class imbalance inherent in intrusion detection datasets. We proposed GTCN-G, a novel hybrid model that synergistically integrates a Gated Temporal Convolutional Network for temporal analysis with an adaptive Graph Convolutional Network for structural feature extraction. The core innovation is a residual learning mechanism within the graph attention component, which is specifically designed to preserve features from minority attack classes. Through extensive experiments on the UNSW-NB15 and ToN-IoT datasets, we demonstrated that GTCN-G achieves state-of-the-art detection performance, significantly outperforming strong GAT and E-GraphSAGE baselines. The findings validate that our fusion of temporal and graph-based learning, coupled with a targeted mechanism to combat class imbalance, is a robust strategy for developing next-generation NIDS. Building on these results, future work will focus on evaluating the model's computational efficiency for real-time deployment, investigating its interpretability to understand its decision-making process, and exploring its generalization to other graph-based security domains like system call analysis or fraud detection.

\vspace{12pt}

\end{document}